# LANGUAGE IDENTIFICATION AS IMPROVEMENT FOR LIP-BASED BIOMETRIC VISUAL SYSTEMS

*Lucia Cascone*⋆    *Michele Nappi*⋆    *Fabio Narducci*⋆

⋆ Department of Computer Science, University of Salerno, Salerno, Italy

## ABSTRACT

Language has always been one of humanity's defining characteristics. Visual Language Identification (VLI) is a relatively new field of research that is complex and largely understudied. In this paper, we present a preliminary study in which we use linguistic information as a soft biometric trait to enhance the performance of a visual (auditory-free) identification system based on lip movement. We report a significant improvement in the identification performance of the proposed visual system as a result of the integration of these data using a score-based fusion strategy. Methods of Deep and Machine Learning are considered and evaluated. To the experimentation purposes, the dataset called laBial Articulation for the proBlem of the spokEn Language rEcognition (BABELE), consisting of 8 different languages, has been created. It includes a collection of different features of which the spoken language represents the most relevant, while each sample is also manually labelled with gender and age of the subjects.

## 1. INTRODUCTION

Lip movement biometrics is an emerging technology that aims to identify individuals by analyzing their unique lip movements during speech, much like fingerprints or DNA structures [1]. This technology has potential applications in forensics and security, particularly in situations where other biometric identifiers are obscured or unavailable. However, lip movement biometrics is still a developing field and faces technical challenges, such as accurately detecting and analyzing lip movements under harsh environmental conditions.

In this paper we investigate the possibility of identifying an individual using only visual information from the lip region and no audio information [2], with recognition purposes in mind and a multilingual context. For the first time, to the best of our knowledge, we fuse lip movement biometric data with the soft biometric information of visual language identification (VLI), which is the process of identifying a speaker's language solely through lip movements, without audio. The latter little-explored field of research could be enhanced precisely by its beneficial use as soft biometrics to improve the performance of recognition systems. Therefore, an ad hoc dataset called BABELE was created to ensure a balanced distribution of subjects for different language classes, age and gender-related demographic factors. The main contributions of this work are the following: 1) The first use of linguistic information as soft biometrics to enhance the performance of an identification system based on lip movement in which the audio component is not utilised. 2) Presentation of a human-labeled public data set containing information on identity, eight languages, gender, and age. In addition to the original full-length videos with audio, there are five 10-second lip-only clips without audio for each identity. Captures are not controlled, but video quality standards are challenging. 3) Benchmark results for VLI task using only lip area as input on a public dataset with 3 different strategies, based on Machine Learning (ML) and Deep Learning (DL) approaches. 4) Comparison of results for the VLI task under independent and subject-dependent conditions.

## 2. RELATED WORKS

Language recognition generally refers to the automatic identification of language from speech input. Automatic language identification from audio speech is a well-studied problem with well-known difficulties. Given these issues and the contributions this technology would make to applications ranging from natural language processing (NLP) to speech synthesis systems and multilingual information retrieval, it is clear that the potential of VLI has not yet been fully realised. Psycholinguistics has studied the ability of humans to recognise languages by observing the lip movements of speakers [3]. Soto et al. demonstrate that facial vocal information alone may suffice for language recognition. Newman and Cox's are among the few articles that have considered the VLI as a classification problem [4], [5]. However, these early studies are not particularly difficult because the videos used are constrained by the study conditions, which include a limited number of subjects reading a given text and an analysis method that does not employ DL techniques. Facial landmark approach in [6] that is limited to a single binary classification between English and French. The most interesting work in this area is that of [7], although it is unclear the input they use to extract lip movements whether that is only the lip area, the whole face or a considerable part. The impact of more information could in fact affect the performance of the model. Problematic to derive such information or do a comparative study since the dataset used has not been released yet.

| Total Whole Videos | | | | | |
|---|---|---|---|---|---|
| | Man | | Woman | | |
| | U 30 | O 30 | U 30 | O 30 | Tot |
| **Italian** | 8 | 6 | 10 | 8 | 32 |
| **English** | 7 | 6 | 7 | 12 | 32 |
| **German** | 8 | 6 | 9 | 9 | 32 |
| **Spanish** | 7 | 11 | 6 | 8 | 32 |
| **Dutch** | 6 | 17 | 5 | 4 | 32 |
| **Russian** | 2 | 14 | 4 | 12 | 32 |
| **Japanese** | 9 | 8 | 6 | 9 | 32 |
| **French** | 8 | 9 | 8 | 7 | 32 |
| | | | | | **256** |

**Table 1**. Dataset Information (under 30, denoted as U 30, vs. over 30, denoted as O 30).

## 3. DATASET DESCRIPTION

In order to effectively identify a subject using only visual information from the lip region, and incorporating language information as a soft biometric, a specific and well-defined dataset is required. This led to the creation of the BABELE dataset for the study of laBial Articulation for the proBlem of the spokEn Language rEcognition. The dataset comprises videos sourced from YouTube, featuring a diverse range of individuals such as journalists, YouTubers, sportsmen, and politicians, among others. The videos were manually labelled according to the language spoken, gender, and age demographic, ensuring a balanced representation across all groups. The languages included in the dataset are French, Italian, English, Spanish, German, Dutch, Japanese, and Russian, with statistics provided in Table 1. The selection criteria for videos included a minimum duration of 5 minutes, single subject focus, frame rate between 25 and 60 frames per second, and resolution of 720p or higher. Once the dataset was organised and balanced, the subvideo extraction phase began, using landmarks to identify lip movement and associating five 10-second subvideos of the lip area with each subject, excluding any other facial features. These sequences were resized to 200 x 300 pixels and centred within the video scene, resulting in a total of 1280 video sequences, comprising eight languages, 256 identities, and five 10-second lip-only videos for each identity. The dataset has been separated into training and testing sets and is publicly available at the provided link: **shorturl.at/AETX6**

## 4. METHODS

The purpose of our study was to investigate the efficacy of lip movement analysis in identifying a subject, while also integrating language information as a soft biometric feature. To this end, we carried out two distinct experiments, one for each task: subject identification and VLI. The results of both experiments were then combined to obtain the final outcome. For the language identification task, we evaluated three different approaches: two based on DL, namely ConvLSTM and BLSTM, and one based on ML, using the SVM algorithm. As for the subject identification task, we opted for the ConvLSTM model, which we deemed the most suitable choice for the purpose, as we will explain in more detail below. The input are the 10-second lip videos [1].

**ConvLSTM.** *Pre-processing.* To optimize the computational resources of the implemented neural models, the videos were converted to grayscale, and the dataset was transformed into binary format to reduce memory usage and accelerate uploads to cloud-computing environments. The serialization and deserialization of objects into a byte stream were carried out using Python's dill library. This technique enables the transmission of an object's complete state so that it can be perfectly reconstructed through deserialization.
*Architecture.* The DL architecture presented is illustrated in Figure 1. The ConvLSTM network merges the advantages of both the convolutional neural network (CNN) and the long-term memory network (LSTM) and possesses strong capability in extracting time series information, making it a suitable solution for effectively addressing these problems.

**BLSTM.** *Pre-processing.* In line with the findings of the study [8], in order to enhance the effectiveness of lip reading, we decided to assess the impact of three different edge detection methods: Sobel [9], Canny [10] and Laplacian [11]. Following a preliminary investigation with a limited number of languages, we selected Sobel as pre-processing technique.
*Architecture.* These models have the potential to capture long-term temporal dependencies, extracting visual features and transforming them into more distinctive representations that are useful for language recognition. We selected this approach based on the insights presented in [7].

**SVM.** *Pre-processing.* MediaPipe [12] extracts eight specific labial landmark coordinates from each video frame. These landmarks are depicted in Figure 1. After selecting one landmark as the pivot, the algorithm calculates the distances between it and the other seven landmarks in each frame. The distances are evaluated using three metrics: Euclidean, Manhattan, and Cosine. The resulting distance values are concatenated to construct the feature vector.
*Architecture.* To find the optimal configuration, an exhaustive search was conducted among a set of hyperparameters. Additionally, a *k*-fold strategy was employed to prevent overfit-ting, with *k* values ranging from 2 to 10. For each exhaustive search, the SVM algorithm was tested using 9 different values of *k*, and the best performing value was selected.

---
[1] The source code of our implementation will be made available upon request to the corresponding author.

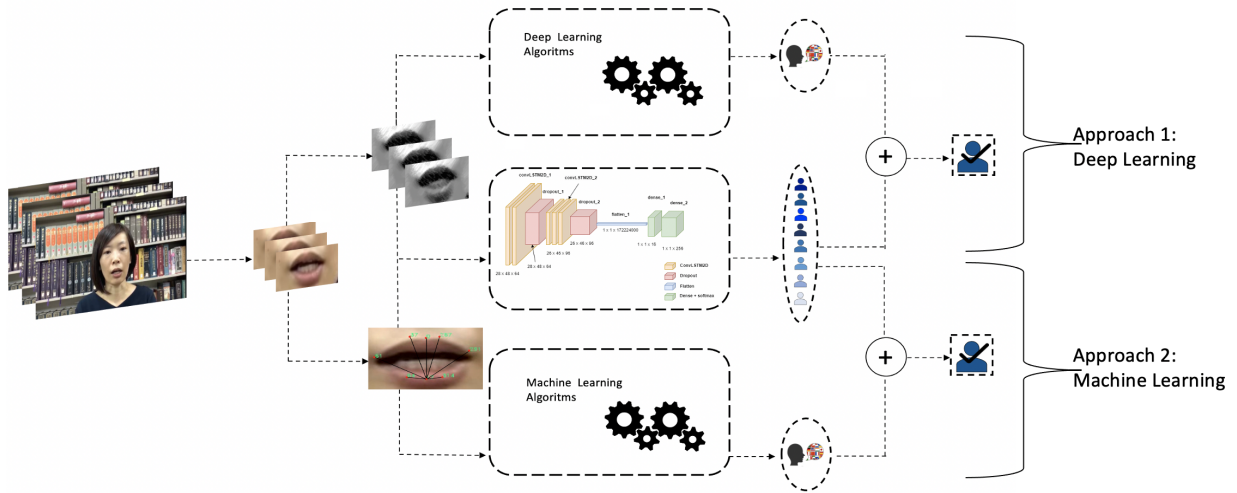

**Fig. 1**. The proposed framework for a lip-based identification in a visual system.

**Fusion module.** *Score-based Fusion Schemes.* Score-level fusion combines the output of matching scores from two classification tasks to determine an individual's identity. This involves generating a single scalar score, which the biometric system uses to make the final determination. From the eight identities deemed most reliable by the identification model, the system selects the identity with the highest score whose language matches the predicted language.

## 5. EXPERIMENTAL SETTING

**Visual Language Identification.** The VLI experimentation is conducted through both subject-independent and subject-dependent procedures. Essentially, the difference lies in how the dataset is partitioned for the evaluation of training and testing. *Subject-dependent protocol.* The dataset was partitioned such that each subject had four videos designated for training and one video designated for testing. Specifically, 80% of the videos were allocated for training data and 20% for testing data. *Subject-independent protocol.* The dataset was partitioned to ensure that videos belonging to the same subject did not appear in both the training and testing sets. To achieve this, four subjects were randomly selected for testing, four for validation, and the remaining subjects for training. Specifically, 75% of the data was allocated for training, while 12,5% each were allocated for testing and validation. The training and validation data have been merged for the ML approach.

**Visual lip-based Identification.** The subject identification

| VLI Model | VLI Accuracy | | Visual lip-based Identification |
|---|---|---|---|
| | Subject Independent | Subject Dependent | ConvLSTM Accuracy |
| ConvLSTM | 36,78% | 74,21% | 49,60% |
| | **Fusion** | 55,85% | |
| BLSTM | 22,50% | 86,32% | 49,60% |
| | **Fusion** | **60,93%** | |
| SVM | 37,50% | 26,87% | 49,60% |
| | **Fusion** | 32,03% | |

**Table 2**. Summary of the results obtained with respect to the two experiments individually and then after merging the relevant information.

task employed the same dataset partitioning as the subject-dependent VLI protocol.

## 6. DISCUSSION

We have tabulated the results of our experiments in Table 2. Let us now examine the outcomes of the two experiments in greater detail.

### 6.1. Visual Language Identification

We show the confusion matrices for the predictions of the model in Figure 2. The results indicate that the biometric component, which includes unique characteristics of an individual beyond just lip movement, such as visual information, has a significant impact on the classification task, even when, as in our case, the area of interest is limited to the lip region. Examining the results in more detail, it is noteworthy that despite the obvious differences between experiments that include or exclude the biometric component, for both models some theoretical considerations are confirmed. For example, for both there is a tendency not to confuse Russian with English (Figure 2, confusion matrices for the ConvLSTM model). Due to the much higher number of open and closed vowels in English than in Russian, English pronunciation differs significantly from that of the latter language. In addition, Russian has many more soft vowels such as "o" and "e", while English has many more hard vowels such as "a" and "e". English has no guttural sounds such as "g" or "k", while Russian also produces these sounds with the lips. In addition, the position of the lips can produce a number of sibilant sounds in Russian. In English these sounds do not exist. In English, changes in lip position during sound production are less pronounced and lip articulation is less noticeable. Because of these variations, the two idioms are very different from each other. Regarding the ML approach (where the results for the best configuration are reported), it can be unequivocally stated that the biometric component does not present a significant discriminative contribution that could enhance the model's performance. Undoubtedly, the larger sample size used for the subject-independent experiment was a crucial factor in improving the performance. However, further investigations are warranted to comprehensively assess this aspect.

### 6.2. Visual lip-based Identification

The aim of the experiment was to identify a person by analyzing their lip movements. Table 2 reports the results of predicting the identity of 256 different subjects using the ConvLSTM model, which achieved an accuracy of approximately 50%. In three separate trials that involved integrating language information through a score-based fusion system, a significant improvement of over 10% was observed when language was predicted using the BLSTM model and about 6% with ConvLSTM. However, the ML approach failed to achieve good interference ability for the VLI task, resulting in a worsening of the results. From the experiments conducted, the results indicate that linguistic identity can be considered as a soft biometric feature that, when properly integrated into the recognition system, can improve its accuracy. It should also be noted that when the two types of information are combined, a considerable percentage of errors in predicted identities are due to errors in the subject identification model. For instance, in many cases where the predicted language was correct, integrating it with identification information did not change the predicted identity because none of the eight possible subject predictions was correct. Figure 3 shows that about 60% of the errors obtained with language integration (ConvLSTM model) were caused by errors in the subject identification model, where the predicted language was correct, but none of the eight most plausible subject identities selected by the model was correct.

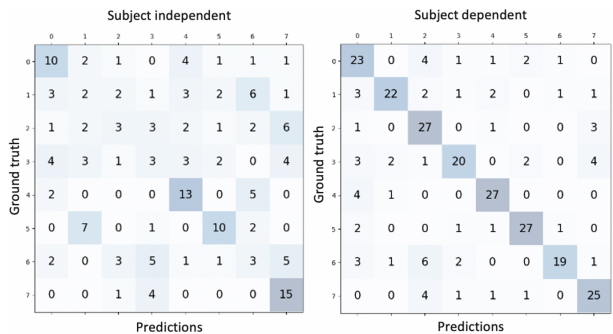

**Fig. 2**. Confusion matrices for the two VLI studies based on the subject-independent and subject-dependent protocol, ConvLSTM model. The language-related codes are as follows. French is 0; Japanese is 1; English is 2; Italian is 3; Dutch is 4; Russian is 5; Spanish is 6; and German is 7.

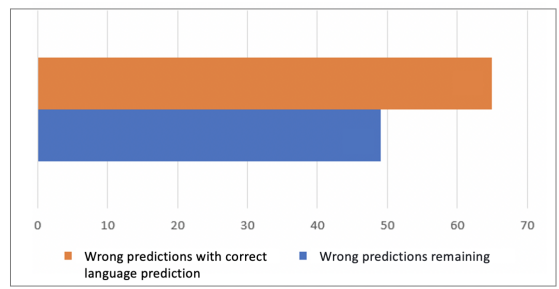

**Fig. 3**. Analysis of incorrect predictions after module fusion.

### 7. CONCLUSIONS

In this work, we aimed to use spoken language information to improve the performance of a visual identification system that only receives lip area video input without audio. The results confirm our hypothesis when the chosen model successfully infers language for prediction. In future experiments, we believe that the contribution of language could have an even greater impact, as there was a significant error in the model's identity predictions that then affected the correct integration of language. As this is still a relatively new and developing field, further research will be necessary to establish its reliability and accuracy for effective application, including as a forensic tool. Overall, the findings demonstrate the potential for language to enhance visual identification systems and open up new possibilities for future research and development.